\documentclass[letterpaper, 10 pt, conference]{ieeeconf}  
\usepackage[left=46pt,top=58pt,right=60pt,bottom=47pt]{geometry}
\usepackage{amsmath}
\usepackage{algorithm,algpseudocode}
\usepackage{graphicx}
\usepackage{float}
\usepackage{amssymb}
\usepackage{amsbsy}
\usepackage{nccmath}
\usepackage{xcolor}
\usepackage{graphics}
\usepackage{cite}
\usepackage{multirow}
\usepackage[colorinlistoftodos,prependcaption,textsize=tiny]{todonotes}
\usepackage{mathtools}

\usepackage{array,booktabs}

\DeclarePairedDelimiter\floor{\lfloor}{\rfloor}

\IEEEoverridecommandlockouts                   
\title{\LARGE \bf
Task Planning on Stochastic Aisle Graphs for Precision Agriculture
}

\author{Xinyue Kan,$^1$ Thomas C. Thayer,$^2$ Stefano Carpin,$^2$ and Konstantinos Karydis$^1$
\thanks{$^1$ Dept. of Electrical and Computer Engineering, University of California, Riverside. 
	Email: {\{xkan001, karydis\}@ucr.edu}.} 
\thanks{$^2$ Dept. of Computer Science and Engineering, University of California, Merced. 
	Email: {\{tthayer, scarpin\}@ucmerced.edu}.}

\thanks{We gratefully acknowledge the support of NSF under grants \#IIS-1724341, \#IIS-1901379 and \#DGE-1633722, and USDA-NIFA under grants \#2017-67021-25925 and \#2021-67022-33453. Any opinions, findings, and conclusions or recommendations expressed in this material are those of the authors and do not necessarily reflect the views of the funding agencies.}%
}

\begin{document}

\maketitle
\thispagestyle{empty}
\pagestyle{empty}

%%%%%%%%%%%%%%%%%%%%%%%%%%%%%%%%%%%%%%%%%%%%%%%
\begin{abstract}
This work addresses task planning under uncertainty for precision agriculture applications whereby task costs are uncertain and the gain of completing a task is proportional to resource consumption (such as water consumption in precision irrigation). The goal is to complete all tasks while prioritizing those that are more urgent, and subject to diverse budget thresholds and stochastic costs for tasks. To describe agriculture-related environments that incorporate stochastic costs to complete tasks, a new Stochastic-Vertex-Cost Aisle Graph (SAG) is introduced. Then, a task allocation algorithm, termed Next-Best-Action Planning (NBA-P), is proposed. NBA-P utilizes the underlying structure enabled by SAG, and tackles the task planning problem by simultaneously determining the optimal tasks to perform and an optimal time to exit (i.e. return to a base station), at run-time. The proposed approach is tested with both simulated data and real-world experimental datasets collected in a commercial vineyard, in both single- and multi-robot scenarios. In all cases, NBA-P outperforms other evaluated methods in terms of return per visited vertex, wasted resources resulting from aborted tasks (i.e. when a budget threshold is exceeded), and total visited vertices. 
\end{abstract}

%%%%%%%%%%%%%%%%%%%%%%%%%%%%%%%%%%%%%%%%%%%%%%%
\section{INTRODUCTION}

Autonomous agricultural mobile robots are becoming increasingly more capable at performing \emph{persistent missions} such as monitoring crop health indices~\cite{Bietresato16} and sampling specimens~\cite{AuatCheein17} across extended spatio-temporal scales to enhance efficiency and productivity in precision agriculture~\cite{roldan2018robots}. 
An autonomous robot (or a team of them) needs to perform certain tasks in distinct locations of the operating environment subject to a specific budget~\cite{bochtis2015route} on the actions the robot can take (e.g., a maximum capacity of soil samples to carry~\cite{vaeljaots2018soil}). During in-field operations, the actual costs to complete tasks can be uncertain whereas expected costs may be known. In addition, some tasks can be more urgent than others, hence they will have to be prioritized. It is often the case~\cite{Pretto_2020,roldan2018robots,bonadies2016survey} that there exists some prior information about a required task (e.g., older measurements of soil moisture~\cite{thayer18}) that can bias robot task assignment(s). Hence, it is necessary to develop approaches that utilize limited prior information to plan tasks with uncertain costs and priority level.

There exist two key challenges for efficient robot task allocation in precision agriculture. 
First, prior maps can indicate biases in task assignments, but may not be trustworthy. This is because conditions in the agricultural field can change rapidly~\cite{gago2015uavs}, are dynamic~\cite{zheng2009spatiotemporal,jones2013plants}, and may be hard to predict ahead of time~\cite{kelly2019challenges}.  
Second, as the budget is being depleted, the robot needs to periodically return to a base station (e.g., to drop collected samples and/or recharge).
Addressing these two challenges simultaneously poses a two-layer intertwined decision making under uncertainty problem: \emph{How to perform optimal sampling given an approximate prior map, and how to decide an optimal stopping time (i.e. to return to base) to avoid exceeding a given task capacity?} This paper introduces a new stochastic task allocation algorithm to balance optimal sampling and optimal stopping when task costs are uncertain.

A direct approach for persistent sampling (and/or monitoring) is to survey the entire space and perform the desired task(s) sequentially~\cite{Hoang2013,kan19,kan20}. 
The main drawback is that the robot would then exhaustively visit all desired sampling locations in the environment without prioritizing locations that would yield a higher gain or would be more time-critical. 
Orienteering~\cite{Thayer18_2, Thayer19,gunawan2016orienteering,Tokekar16} can address part of this drawback by determining paths that maximize the cumulative gain under a constant budget. 
The robot prioritizes visiting adjacent locations if they jointly yield higher gains than isolated high-gain locations, and provided that any budget constraints are not violated~\cite{Thayer18_2,Thayer19}. However, this strategy can be insufficient for missions where some tasks are more urgent than others. 
For instance, several existing robot task allocation strategies, albeit for distinct application domains~\cite{auat2017agricultural,wurman2008,veloso2015,khaluf2019local}, typically consider a deadline~\cite{Ma2018} or user-defined importance levels. 
In precision agriculture, overhead imagery (e.g., thermal imaging) can help pinpoint locations that appear to be under water stress~\cite{gago2015uavs}, in which case sampling leaves or soil in those areas should be prioritized.   
We formalize the notion of tasks with distinct urgency (e.g., a closer deadline or greater importance) by assigning a \emph{priority level}~\cite{becker2005robust,Maoudj16} to tasks.

Besides the task priority level, deciding a next task for a robot to complete is also dependent on available budget, which can be of multiple types. For instance, the number of locations that a robot can visit and sample from in one `trip' is constrained by both the energy capacity to move between locations and the robot's sample payload capacity. Exceeding the energy budget can prevent the robot from returning to the base station to recharge and drop collected samples, whereas exceeding the sample payload capacity may cause potential robot and sample damage. 
Here we consider an \emph{energy budget} for the robot moving between locations, and a \emph{resource budget} linked to task execution. 
The two budgets are independent of each other, and both can be reset to their initial values when the robot returns to the base station. The actual amount of resources consumed to execute a task can differ from what is the expectation in practice. In fact, the actual amount of resources consumed for task execution is revealed only after the task has been completed. 
To model this, we consider the cost to complete a task to be a stochastic random variable that follows some known distribution. The cost to move between locations, however, is considered to be deterministic~\cite{thayer18}. Specific details are given in the following.

This paper introduces a new stochastic task allocation approach, termed Next-Best-Action Planning (NBA-P), for task planning under uncertainty in precision agriculture.  The paper also contributes a new Stochastic-Vertex-Cost Aisle Graph (SAG). SAG is an extension of the aisle graph~\cite{thayer18,Sorbelli20}, which is often used to describe agriculture-related environments. The main novelty of SAG is that it can represent uncertain task costs. 
Using SAG, our proposed NBA-P algorithm simultaneously determines 1) how to optimally schedule which tasks to perform at run-time, and 2) when to optimally stop performing new tasks and return back to the base station also at run-time.  NBA-P ensures that urgent tasks are prioritized subject to both energy and resource budgets. In addition, it can be extended to multi-robot teams. We test our method in single- and multi-robot scenarios using both simulated data and 10 real-world datasets collected in a commercial vineyard at central California. In all cases, NBA-P achieves higher efficiency than naive lawnmower, informed lawnmower, and series Greedy Partial Row planners~\cite{zelinsky1993planning,hameed2013optimized,oksanen2009coverage} in terms of more return per visited vertices,
less resources wasted because of aborted tasks, and less total visited vertices.

\section{Related Work}

\emph{Aisle graphs}~\cite{thayer18,Sorbelli20} can model motion constraints emerging when robots navigate in structured environments such as agricultural fields. Vertices denote possible task locations, and edges represent connections between locations. Any two rows connect to each other only via the two end vertices. 
Moving backwards is not allowed. Hence, if a robot enters a row, it will have to reach the row's other end then to move to any other row. In the original aisle graph formulation~\cite{thayer18,Sorbelli20}, vertices and edges are associated with known and constant reward and movement costs, respectively. Our proposed extension, SAG, can also represent uncertain task costs.

\emph{Orienteering} can tackle persistent sampling on aisle graphs. Even with motion constraints introduced via aisle graphs, orienteering remains an NP-hard problem and thus greedy heuristics are often employed~\cite{thayer18}. 
Recent efforts on stochastic orienteering associate stochastic costs to graph edges and propose a time-aware policy for a robot to adjust its path to avoid exceeding a certain budget~\cite{Thayer20}.  However, addressing cases that involve uncertain task cost on vertices for aisle graphs remains open. Our proposed NBA-P tackles the problem by simultaneously considering uncertain task costs on vertices and deterministic costs on edges.

The optimal stopping framework~\cite{shiryaev2007optimal} can be used to investigate the (optimal) criteria to terminate a process while incorporating uncertainty~\cite{Chung12}. In most cases~\cite{abdelaziz2007optimal,peskir2006optimal,chen2014adaptive}, data arrive in sequence, and irrevocable decision has to be made as to when the expected return is maximized. Optimal stopping has been used in robotics applications like target tracking~\cite{Best17} and marine ecosystem monitoring~\cite{das2015data}. However, no motion constraints, like those imposed by aisle graphs, apply to robot actions, and hence existing methods cannot be ported over to operations on aisle graphs. Paths planned with NBA-P fill the gap, as they directly apply to environments with motion constraints captured by aisle graphs.

The proposed method applies when: 1) the motion constraints in the application environment can be captured by a SAG; 2) the cost of completing tasks follow exponential distributions; and 3) the obtained gain by completing a task is proportional to the actual task cost.

\section{Stochastic Task Allocation Problem Setup}

We first define the \emph{Stochastic-Vertex-Cost Aisle Graph (SAG)}, to incorporate uncertain task cost on vertices. Then, we present this paper's problem setup utilizing SAG.

\subsection{The Stochastic-Vertex-Cost Aisle Graph (SAG)}
We propose SAG as a way to extend the original aisle graph~\cite{thayer18,Sorbelli20} to handle missions consisting of tasks with priority levels and stochastic execution costs. There are three main differences between SAG and the original aisle graphs. 1) SAG considers stochastic costs for task execution at vertices. 2) Vertices in SAG are associated with task priority levels. 3) The gain, which describes the benefit of completing a task, is proportional to the actual resource consumption if the task is fully completed. With more resource consumption, higher gain could be obtained, e.g., higher quality information during soil sampling process, or better field hydration/irrigation results.
Note that no gain will be obtained if 1) the resource budget is exceeded during task execution, and the task is aborted, or 2) a robot only passes through a vertex on its way without performing a task.
In contrast, in the original aisle graph rewards are constant and can be collected immediately when passing through vertices. 

\begin{figure}[!h]
\vspace{-11pt}
    \centering
    \includegraphics[width=0.45\linewidth]{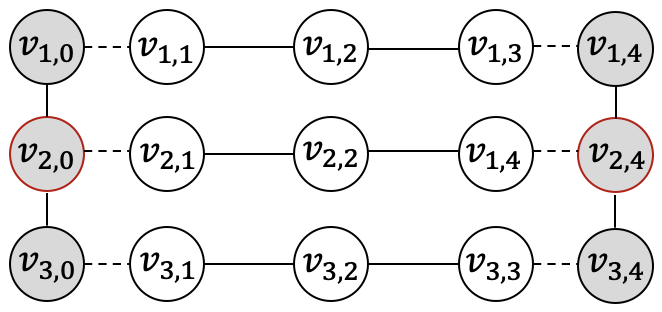}
    \vspace{-6pt}
    \caption{SAG $A_s(3,3+2)$ with 3 rows and 5 columns. Grey nodes are end vertices to connect rows. Vertices with red edges are base stations.}    \label{fig:2}
    \vspace{-7pt}
\end{figure}

Given a field that contains $m$ rows and $n$ columns (where $n$ denotes the total number of possible sampling locations in each row), its SAG representation is an (undirected) graph $A_s(m,n+2) = (V, E)$ where $V$ and $E$ are the sets of vertices and edges, respectively. Note that in the graph representation we add two additional `virtual' columns at indices $j=0$ and $j=n+1$ that connect the $m$ rows; virtual vertices carry no gain. An example of a $A_s(3,5)$ graph is given in Fig.~\ref{fig:2}. 

The set of edges $E$ is built as follows:\footnote{For clarity and completeness, we follow and partially adapt the definition of the original aisle graphs from~\cite{thayer18,Sorbelli20}.}
\begin{itemize}
    \item Vertices $v_{i,j}$ with $i\in [1,m]$ and $j\in[1,n]$ have two edges, $e_{i,(j-1)^+}$ and $e_{i,j^+}$.
    \item Vertices $v_{i,j}$ with $i \in (1,m)$ and $j \in \{0, n+1\}$ have three edges: if $j = 0$ then $\{e_{i,0^+}, e_{(i-1)^+,0}, e_{i^+,0}\}$ and if $j = n+1$ then $\{e_{i,n^+}, e_{(i-1)^+,n+1}, e_{i^+,n+1}\}$.
    \item The four corner vertices are $v_{1,0}, v_{m,0}, v_{1,n+1}, v_{m,n+1}$, each of which has two edges.
\end{itemize}

Set $S$ contains all priority levels in $A_s$. Let $c_v: V \xrightarrow{} \mathbb{R}_{\geq 0}$ and $c_e: E \xrightarrow{} \mathbb{R}_{\geq 0}$ be the costs for task execution at vertices and movement on edges, respectively. The actual resource consumption to complete a task at vertex $v\in V$ follows an exponential distribution, $c_v(v)\sim Exp(\bar{w}_s)$, where $\bar{w}_s$ is the mean cost of all tasks with priority level $s \in S$. 
The actual task cost is not known before task completion, and is independent between tasks at different locations. The cost of movement on edges is a known constant. Function $f:V \xrightarrow{} S$ returns the priority level of a vertex, and $f(v_{i,j}) = 0, v_{i,j}\in V$ indicates no tasks at a vertex.\footnote{For clarity, we will henceforth write $f(v_{i,j})$ as $f_{(i,j)}$. Any other functions that take vertices as input will be shortened similarly.}  Condition $f_{(i_1,j_1)} < f_{(i_2,j_2)}$ implies that the task at $v_{i_2,j_2}$ is more urgent than the task at $v_{i_1,j_1}$. In other words, %to prioritize tasks with higher priority level, it is reasonable to assume that 
if the same amount of resources is consumed at $v_{i_1,j_1}$ and $v_{i_2,j_2}$, higher gain is obtained at $v_{i_2,j_2}$. Once a task is completed, its priority level is set to $0$. 

Let $r: V \xrightarrow{} \mathbb{R}_{\geq 0}$ be the actual gain obtained when completing a task. Function $\mu:S\xrightarrow{} \mathbb{R}_{>0}$ maps each priority level to a deterministic positive value, which indicates the gain-to-cost ratio of completing a task of given priority level. Then, $r_{(i_1,j_1)}= \mu(f_{(i_1, j_1)})c_{v(i_1, j_1)}$, and $\mu(f_{(i_1, j_1)}) < \mu(f_{(i_2, j_2)})$ if $f_{(i_1,j_1)} < f_{(i_2,j_2)}$ for $v_{i_1, j_1}, v_{i_2, j_2} \in V$. Vertices $v_{i,j}$ with $i \in [1,m]$ and $j = n+1$ are virtual nodes to connect rows, hence $f_{(i,j)} = 0$, $c_{v(i,j)}=0$, $r_{(i,j)}=0$. Edges $e_{i,0^+}$ and $e_{i,n^+}$ with $i \in [1,m]$ has $c_e(e_{i,0^+}) = c_e(e_{i,n^+}) = 0$.

%\paragraph*{Discussion} 
Priority levels can be user-defined or estimated via any prior environment %reward 
maps. 
The latter can be determined based on collected data, e.g., difference between ideal and sampled soil moisture levels~\cite{thayer18}. However, prior information may be approximate and thus lead to suboptimality if directly set as priority levels for vertices.
A way to assign priority levels from prior information is to set thresholds so that data within a range yield the same priority level. Only same types of tasks with same expected cost can be set at same priority level.

\subsection{Stochastic Task Allocation on SAGs}

A mission on SAG $A_s(m,n+2) = (V,E)$ comprises tasks located at $v\in V_T \subset V$. Given energy budget for moving along edges and resource budget for executing tasks on vertices, to complete all tasks in $V_T$ so that:
\begin{itemize}
    \item C1: Tasks are prioritized according to priority level.
    \item C2: The number of tasks being aborted because of exceeding the resource budget (at run-time) are minimized.
\end{itemize}

C1 enforces the time-critical decision making, whereas C2 ensures efficiency of mission completion. When a task is aborted, no gain is obtained and both consumed resources and energy spent moving to that vertex are wasted. Aborting tasks will also cause delays on mission completion time. 
To avoid exceeding the resource budget at run-time, the robot thus needs to determine an optimal stopping time. 
Its next action should be to either 1) perform another feasible task of the highest possible priority level (which we describe how to set next), or 2) stop performing tasks and return to the base station. Since the actual cost is unknown before completing a task, the next action and corresponding paths are determined in an adaptive manner based on remaining budget at run-time. 

\section{Proposed Task Planning Algorithm}
Our proposed Next-Best-Action Planning (NBA-P) approach balances sampling feasible vertices on SAG and determining when it is preferable to exit (i.e. return to base station) based on remaining resource and energy budgets. 
When sampling feasible vertices, we use a three-phase approach. \underline{Phase 1}: sample feasible vertices subject to resource budget; \underline{Phase 2}: sample feasible vertices from phase 1 subject to energy budget; \underline{Phase 3}: select a row to proceed and plan corresponding paths. 
When sampling in phase 1, we start from the highest priority level that currently exists. If either phase 1 or phase 2 returns no feasible vertex, we decrease the examined priority level until either feasible vertices are found, or the examined priority level reaches 0, in which case it is optimal to exit. This strategy ensures that tasks with higher priority level are prioritized when possible. 

\subsection{Phase 1: Feasible Vertices Subject to Resource Budget}
To tackle the stochastic task cost, we formulate the next-task selection subject to resource budget as an optimal stopping problem. We employ a one-stage-look-ahead rule: if it is better to \emph{return to base station directly} than to \textit{perform one more task of any priority level then return}, then return at current time. In this phase, we do not need to consider the actual robot position.  
Let $p$ and $q$ be the remaining resource budget and the total gain in the current `trip' (i.e. operation since last visit to a base station), respectively.\footnote{Gain $q$ are set to 0 when the robot resets at the base station.} If a task of priority level $s\in S$ consumes $x$ amount of resources, the return is $\mu(s)x$. Then, in a dynamic programming framework, with $(p, q)$ the state, the expected return function, $\Phi(p, q)$, is
\begin{equation}\label{eq1}
\begin{medsize}
\hspace{-6pt}\Phi(p, q) = 
\begin{cases}
    \max\limits_{s \in S} \big\{ \int_{0}^{p}\lambda_s e^{-\lambda_s x}\Phi(p-x,q+\mu(s) x) dx \big\}  \text{, \hspace{-2pt}if p $>$ 0}\\
    q \text{,  otherwise}\text{,}
\end{cases}\hspace{-16pt}
\end{medsize}
\end{equation}
where $\lambda_s = \frac{1}{\bar{w}_s}$. When $p>0$ (i.e. some resource is available), a task of priority level $s\in S$ which maximizes the return is selected. Otherwise, no task can be completed and the total return remains the same as $q$.

\subsubsection{Single Priority Level for All Tasks}\label{phase1}
We start with the case that all tasks in the mission have the same priority level, $|S|=1$. In this case, we only need to determine the optimal time to exit. According to~\eqref{eq1}, for $s\in S$, the state $(p', q')$ is on the optimal stopping boundary if
\begin{equation}\label{eq2}
q' = \int_{0}^{p'}\lambda_s e^{-\lambda_s x}(q'+\mu(s) x) dx,
\end{equation}
since continuing to perform another task will not result in higher expected return. Hence, the robot should exit if the current state $(p,q)$ satisfies $p<p'$ and $q>q'$, i.e. all tasks are infeasible. Solving~\eqref{eq2} leads to
\begin{equation}\label{eq3}
    q' = \frac{\mu(s)}{\lambda_s}(e^{\lambda_s p'}-1-\lambda_s p').
\end{equation}
Defining function $g:(\mathbb{R}_{\ge 0},S)\rightarrow\mathbb{R}_{\ge 0},(p',s) \mapsto q'$ based on~\eqref{eq3} represents the optimal stopping boundary curve for a given priority level. 
Thus, it is optimal to exit at state $(p',q')$ when $q' \geq g(p',s)$ given a priority level $s\in A_s$. 

\noindent \textbf{Definition 1.} A task of priority level $s$ is \textit{feasible} for the current state $(p,q)$, if $(p,q)$ lies below the optimal stopping boundary curve $g(p',s)$ (Fig.~\ref{fig:3}).

\subsubsection{Multiple Priority Levels Across Tasks} 
If $|S|>1$, the robot determines the candidates with highest possible priority level allowed by the remaining budget. The optimal strategy is to examine the feasibility to perform a task of priority level $s = max(S)$, and then decrease $s$ until a feasible task is found. If no feasible task exists until $s = 0$, then the optimal decision is to return back to the base station.

When multiple priority levels exist, it is not always true that tasks with higher priority levels must be performed before any lower priority rank tasks. To maximize the expected return in one `trip' (i.e. between two times that a robot visits the base station), when the remaining resource budget is not enough for high priority level tasks, a task with lower priority level can potentially be selected to be performed next. However, in some scenarios, a lower priority task will never be selected prior to a higher priority task.

\noindent \textbf{Lemma 1.} At state $(p,q)$, given that tasks with priority level $s \in S$ are infeasible, then all tasks with $s'\in S$ and $s' < s$ must be infeasible if $\bar{w}_{s'} \geq \bar{w}_{s}$.

\begin{figure}[!t]
\vspace{6pt}
    \centering
    \includegraphics[trim={0 0cm 0 0},clip,width=0.95\linewidth]{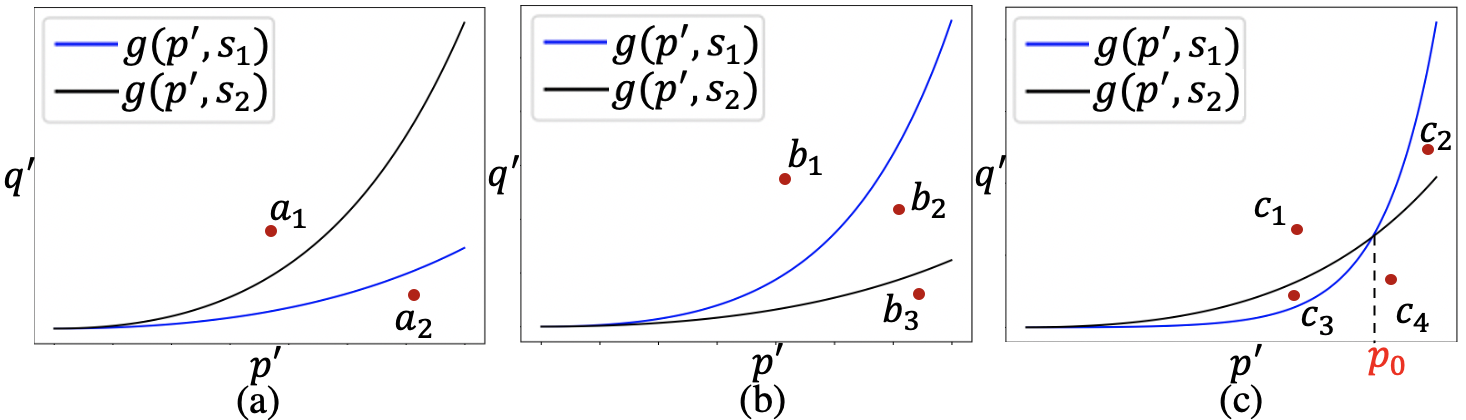}
    \vspace{-8pt}
    \caption{Illustrations of cases described in (a) Lemma 1, (b) Condition 1 and (c) Condition 2. Red points visualize sample states $(p, q)$. In (a)-(c), tasks of priority levels $s_1$ or $s_2$ are feasible if the current state $(p, q)$ is below the curves $g(p',s_1)$ or $g(p',s_2)$, respectively.}
    \label{fig:3}
    \vspace{-3pt}
\end{figure}

\noindent \textbf{Proof.} Let $s_1, s_2 \in S$ : $1\leq s_1 < s_2 \leq max(S)$ and $\mu(s_1) < \mu(s_2)$. The mean costs of $s_1$ and $s_2$ tasks are $\bar{w}_{s_1}$ and $\bar{w}_{s_2}$, respectively. If $\bar{w}_{s_1} \geq \bar{w}_{s_2}$, for any $p > 0$, $g(p,s_2) > g(p,s_1)$.  Hence, if a $s_2$ priority level task is infeasible at state $(p,q)$, i.e. $q \geq g(p,s_2)$, then $q \geq g(p,s_2) > g(p,s_1)$, and
$s_1$ tasks are infeasible too (Fig.~\ref{fig:3}(a)).\hfill~$\blacksquare$

On the other hand, as shown in Fig.~\ref{fig:3}(b)(c), for $s_1, s_2 \in S$ such that $1\leq s_1 < s_2 \leq max(S)$, if $\bar{w}_{s_1} < \bar{w}_{s_2}$, the relationship between boundary curves $g(p, s_1)$ and $g(p, s_2)$ is either Condition 1 (Fig.~\ref{fig:3}(b)) or Condition 2 (Fig.~\ref{fig:3}(c)).
\begin{itemize}
    \item Condition 1 :
\begin{equation}
    g(p, s_2) < g(p, s_1), \forall p > 0,
\end{equation}
\item Condition 2: $\exists p_0 > 0$, such that
\begin{equation}
    g(p,s_2)\begin{cases}
        \geq g(p, s_1), \text{ if } 0<p \leq p_0\\
        < g(p, s_1), \text{ if } p > p_0
    \end{cases}
\end{equation}
\end{itemize}

For Condition 1, a state $(p,q)$ above the curve $g(p,s_2)$ can be still below the curve $g(p,s_1)$, e.g., point $b_2$ in Fig.~\ref{fig:3}(b). In this case, $s_1$ tasks should be performed next even if there still exist $s_2$ tasks. 
For Condition 2, when $p > p_0$, the situation is the same as described above for Condition 1. When $ 0< p \leq p_0$ we reduce to the conditions of Lemma 1, in which case given that $s_2$ tasks are infeasible, $s_1$ tasks must be infeasible (an example is point $c_1$ in Fig.~\ref{fig:3}(c)). 

%%%%%%%%%%%%%%%%%%%%%%%%%%%%%%%%%%%%%
%%%%%%%%%%%%%%%%%%%%%%%%%%%%%%%%%%%%%

Let $Q_1$ be the set containing all feasible vertices subject to a given resource budget. We propose Algorithm~\ref{al:q1} to determine $Q_1$ at a state $(p,q)$. If $Q_1 = \emptyset$, the robot returns to the base station. Otherwise, all vertices in $Q_1$ will continue to be examined in Phase 2 subject to a given energy budget.

\setlength{\textfloatsep}{6pt}%
\begin{algorithm}[h!]
\caption{SampleQone($(p,q), s$)}\label{al:q1}
\begin{algorithmic}[1]
\Procedure{Determine $Q_1$ at state $(p,q)$}{}
\State $s \gets min(s, max(S))$,
$Q_1 \gets \emptyset$
\While{$s>0 \And Q_1=\emptyset$}
\If{$q<g(p,s)$}
 \For{$v\in V_T$}
  \If{$f(v) = s$} $Q_1 \gets Q_1\cup \{v\}$
  \EndIf
\EndFor
\Else
\While{$w_{s-1} \geq w_s$}
 $s\gets s-1$
 \EndWhile
 \State $s\gets s-1$
\EndIf
\EndWhile
\State \Return $Q_1,s$
\EndProcedure
\end{algorithmic}
\end{algorithm}

\subsection{Phase 2: Feasible Vertices Subject to Energy Budget}
From $Q_1$, we continue sampling vertices that satisfy the energy budget constraint. 
Suppose a robot is at vertex $v_{i_c,j_c} \in V$, and a vertex $v_{i,j} \in Q_1$ is a candidate to be examined. Without loss of generality, suppose two base stations are located on row $i_d$, each at one of the end vertices $v_{i_d,0}$ and $v_{i_d,n+1}$. The robot can reset at either one. Vertex $v_{i',j'}$ is feasible if the current remaining energy budget $T$ allows the robot to move to $v_{i',j'}$ then to any base station. Let $t_{\alpha}(i')$ be the cost to move from $v_{i_c,j_c}$ to an end node--either on column $0$ or $n+1$ depending on the robot's moving direction in current row $i'$ (recall backward motion is not allowed). Let $t_{\beta}(i')$ be the cost to move between two end vertices $v_{i',0}$ and $v_{i',n+1}$ in row $i'$, and $t_{\gamma}$ be the cost to move from the end vertex on row $i'$ closest to the robot along its direction of motion to the closest base station. Then, $v_{i',j'}$ is feasible if \footnote{Expressions to compute $t_{\alpha}$, $t_{\beta}$ and $t_{\gamma}$ are given in the Appendix.}
\begin{equation}\label{eq:6}
    T \geq t_{\alpha}(i') + t_{\beta}(i') + t_{\gamma}(i').
\end{equation}
If $v_{i',j'}$ can be reached, all vertices on row $i'$ must be reachable, since $t_{\alpha}(i') + t_{\beta}(i') + t_{\gamma}(i')$ only depends on row $i'$. 
Costs $t_{\beta}$ and $t_{\gamma}$ are fixed for each row and can be precomputed prior to deployment. Cost $t_{\alpha}$ is computed at run-time. 
The set containing all vertices that satisfy both budgets is $Q_2 = \{v_{i',j'} \in Q_1| t_{\alpha}(i') + t_{\beta}(i') + t_{\gamma}(i') \leq T\}$.

\subsection{Phase 3 and Proposed Algorithm}
Phase 3 can be reached if $Q_2 \neq \emptyset$. Note that all tasks at vertices in $Q_2$ have the same priority level $s$ and hence the same expected cost $\bar{w}_s$. Therefore, sampling the next vertex turns into selecting a row $i$ which consists of one or more tasks of priority level $s$. Then, the robot will perform the first encountered feasible task while moving along row $i$. 

Suppose the robot is currently at $v_{i_c,j_c}$ with state $(p,q)$. The row $i\in [1,m]$ is selected such that $\forall i'\in [1,m], i'\neq i_c$,
\begin{equation}\label{eq7}
\begin{medsize}
    \min\{|Q_2(i)|, \floor*{\frac{q}{\bar{w}_s}}\} > \min\{|Q_2(i')|, \floor*{\frac{q}{\bar{w}_s}}\},
\end{medsize}
\end{equation}
\begin{equation}\label{eq8}
\begin{medsize}
    t_{\alpha}(i) \leq t_{\alpha}(i'),
    \end{medsize}
\end{equation}
where $Q_2(i) = \{v_{i',j'}\in Q_2|i'=i \}$, and $\bar{w}_s$ is the mean cost of feasible tasks. By~\eqref{eq7}, the robot is expected to complete more tasks in row $i$ than any other row. 
Thus, row $i$ should be the row that contains the largest number of feasible tasks permitted by remaining budget $q$, according to the expected cost $\bar{w}_s$. If multiple rows return a tie, then the row closest to the robot's current position will be selected as per~\eqref{eq8}.

The proposed Next-Best-Action Planning (NBA-P) approach is formalized in Algorithm~\ref{al:q2}.
NBA-P can be extended to apply to multi-robot teams by sequentially determining the next best action for each robot. In multi-robot implementation, each robot runs NBA-P independently and in parallel, and exchanges information only about the row it currently occupies. For each robot, $Q_2$ has to be modified by removing all vertices in those rows that are occupied by other robots. Note that similar to~\cite{Thayer18_2}, multiple robots can travel simultaneously along the vertical columns $0$ and $n+1$, since space on the boundary of a field is typically much larger.

\vspace{-3pt}
\setlength{\textfloatsep}{0pt}
\begin{algorithm}[h!]
\caption{NBA-P}\label{al:q2}
\begin{algorithmic}[1]
\Procedure{Determine next action at state $(p,q)$}{}
\State $s \gets max(S)$,
$Q_1 \gets \emptyset$, $Q_2 \gets \emptyset$
\While{$Q_2=\emptyset$}
\If{s=0} exit, return to base station
\EndIf
\State $Q_1,s \gets \text{SampleQone}((p,q), s)$
\If{$Q_1 = \emptyset$} exit, return to base station
\Else  
\State obtain $Q_2$ from Eq.~\eqref{eq:6}, $s\gets s-1$
\EndIf
\EndWhile
\State continue with Phase 3
\EndProcedure
\end{algorithmic}
\end{algorithm}
\vspace{-6pt}

\section{Results}
To study the efficiency and effectiveness of the proposed approach, we test with 1) simulated data in a 2-robot scenario, and 2) data collected from a real-world vineyard in 1- and 5-robot scenarios. Testing with simulated data enables parameter tuning so as to study the properties of NBA-P, whereas testing with real-world data reveals the spatial pattern of real tasks that exist in agricultural fields. In both cases, NBA-P is compared against lawnmower planner~\cite{zelinsky1993planning}, which is often seen in agriculture-related applications~\cite{hameed2013optimized,oksanen2009coverage}. 
In naive lawnmower (N-LM), a robot follows meandering paths to survey rows in sequence. When no budget constraint is considered, and when departing from a corner in a square environment, lawnmower will generate the shortest path to survey the entire field in the sense that each vertex is visited only once. In experiments using real-world datasets, NBA-P is also compared against informed lawnmower (I-LM) and Series Greedy Partial Row (S-GPR)~\cite{Thayer18_2}. I-LM attempts to complete a task if the remaining budget is greater than the expected cost of a task. S-GPR is modified to use both energy and resource budgets. In multi-robot cases, each robot runs N-LM, I-LM, and NBA-P independently and in parallel to each other; in S-GPR robots plan their trajectories sequentially. All vertices in occupied rows turn infeasible for other robots so that each row has one robot performing tasks.

\subsection{Testing with Simulated Data and Parametric Study}

We consider a simulated environment $A_s(20,17)$ of 20 rows and 17 columns (including the two virtual columns).  Base stations are located at $v_{10,0}$ and $v_{10,16}$. The cost to move on each edge is 1. Consider two robots deployed from base station $v_{10,0}$ to complete all tasks. Each robot departs with energy budget 80, and resource budget 40.\footnote{Values for both budgets are selected randomly for evaluation purposes. Expected task costs are selected to achieve a budget over expected cost ratio of 20, which helps reveal general patterns of the method.} A vertex is considered ``visited'' if the robot stops at the vertex and attempts to perform a task, regardless whether the task is ultimately completed or aborted. If the resource budget is exceeded before task completion, the task will be aborted, the resources already consumed for this task are considered to be ``wasted,'' and no gain will be obtained. The total gain will be the sum of the actual gain, which is proportional to the actual task cost, at all task-completed vertices. Two cases are studied, 1) $S = \{1\}$, i.e. all tasks have equal priority level, and 2) $S=\{1,2\}$, i.e. two priority levels exist, hence tasks with $s=2$ will be prioritized. In case 1, $\mu(1) = 1, \bar{w}_1 = 2$; and in case 2, $\mu(1) = 1, \mu(2) = 2, \bar{w}_1 = 1.5, \bar{w}_2 = 2$. For each case study, 10 trials are conducted. In each trial, 225 tasks are randomly assigned to 225 vertices in $A_s$, with randomly generated task location and actual task cost. In each trial, the proposed method and the N-LM method are tested on the same simulated environment. 

%%%%%%%%%%%%%%%%%%%%%%%%%%%%%%%%%%%
\begin{figure}[!h]
\vspace{-6pt}
    \centering
    \includegraphics[width =0.40 \textwidth]{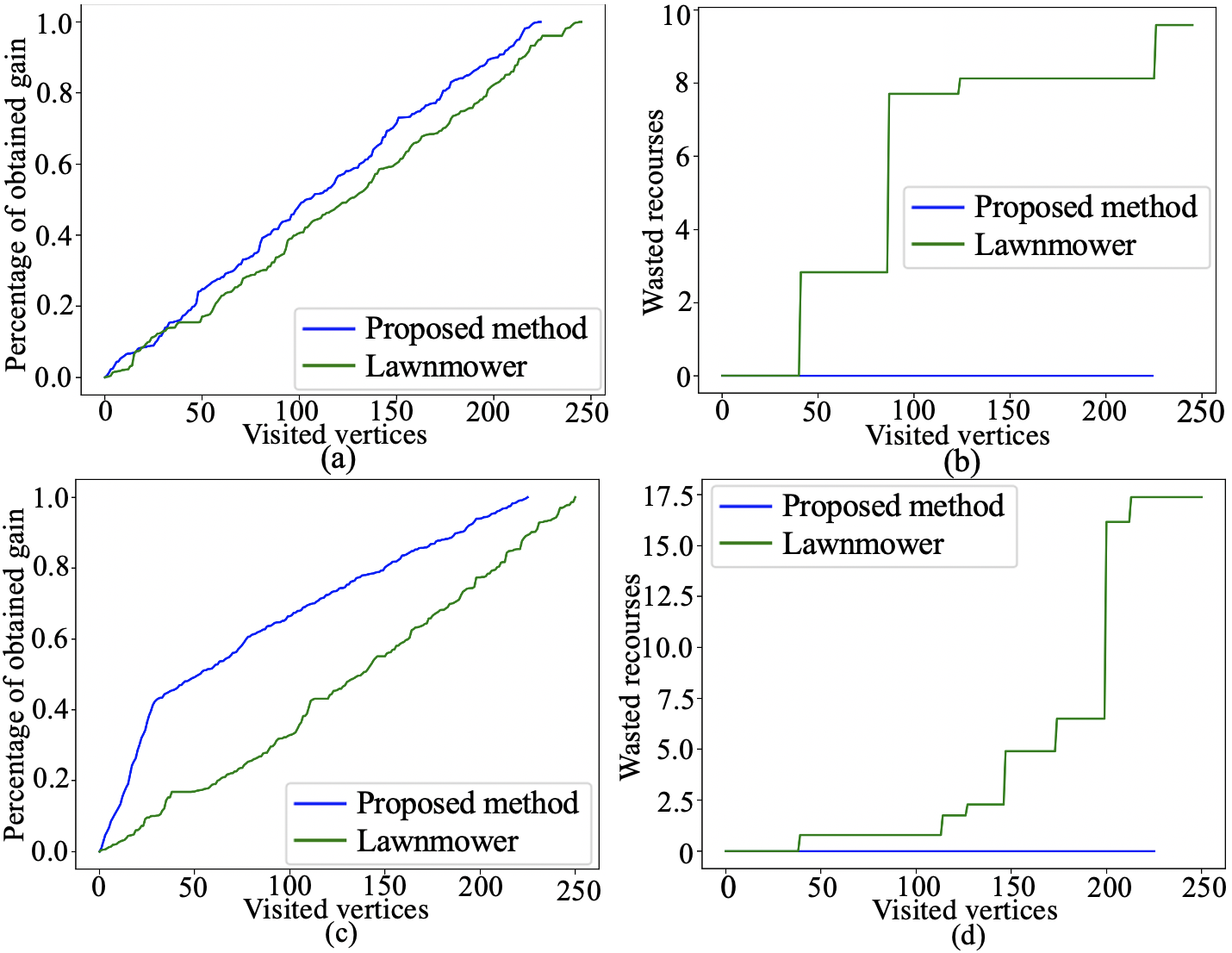}
    \vspace{-9pt}
    \caption{Example of (a) percentage of obtained gain and (b) wasted resources over visited vertices when $|S|=1$. Panels (c) and (d) contain the same information when $|S|=2$.}
    \label{fig:4}
    \vspace{-9pt}
\end{figure}
%%%%%%%%%%%%%%%%%%%%%%%%%%%%%%%%%%%%

Figure~\ref{fig:4} illustrates shows from a sample trial when $|S|=1$ (top) and $|S|=2$ (bottom). Figures~\ref{fig:4}(a) and (c) show the percentage of obtained gain over ground truth total gain as a function of visited vertices (shortened as $r/v$ ratio). Figures~\ref{fig:4}(b) and (d) show the total wasted resources because of aborted tasks as a function of visited vertices (shortened as $w/v$ ratio). Total wasted resources are the sum of resource consumption for all aborted tasks. Total gain, total wasted resources and visited vertices correspond to the sum of those values from both robots. Results suggest that all tasks are completed, and the robots return to the base station.

Table~\ref{table:1} contains the mean and one standard deviation of $r/v$ ratio, $w/v$ ratio, and total visited vertices over 10 trials. Larger $r/v$ suggests higher efficiency since more gain is obtained by visiting the same number of vertices, i.e. same number of attempts to execute tasks. Lower $w/v$ indicates lower rates of aborted tasks, i.e. less resources are wasted by visiting the same number of vertices. Higher $r/v$, lower $w/v$, and less total visited vertices are desired, and these conditions together indicate higher overall effectiveness.

\vspace{-2pt}
\begin{table}[H]
\vspace{-2pt}
\centering
\caption{Results for simulated data over 10 trials}
\label{table:1}
\vspace{-6pt}
\begin{tabular}{cc|c|c|c}
\hline
 &  & r/v ratio& w/v ratio& vertices\\ 
 &  & $(10^{-3})$ & $(10^{-2})$& visited \\ 
 \hline
\multicolumn{1}{c|}{\multirow{2}{*}{$|S|=1$}} & NBA-P & $\pmb{4.42\pm 0.02}$ &$\pmb{1.12 \pm 1.08}$ & $\pmb{226.2\pm 1.1}$\\ \cline{2-5} 
\multicolumn{1}{c|}{} & N-LM & $3.46 \pm 0.18$ & $9.13\pm 1.74$ & $289.4\pm 14.9$ \\ \hline
\multicolumn{1}{c|}{\multirow{2}{*}{$|S|=2$}} & NBA-P & $\pmb{4.44 \pm 0.02}$ & $\pmb{0.33\pm 0.73}$ & $\pmb{225.4 \pm 0.7}$\\ \cline{2-5} 
\multicolumn{1}{c|}{} & N-LM & $3.76\pm 0.1$ & $4.5\pm 1.63$ & $266.2 \pm 7.0$\\
\hline
\end{tabular}
\vspace{-7pt}
\end{table}

Results in Fig.~\ref{fig:4} and Table~\ref{table:1} suggest that, in both cases, NBA-P achieves higher $r/v$ ratio, lower $w/v$ ratio, and less total visited vertices than N-LM. When $|S|=1$, since all tasks have the same priority level, the higher $r/v$ ratio of NBA-P is mainly due to the optimal stopping strategy that helps prevent aborting tasks. When $|S|=2$, the higher $r/v$ is due to both the optimal stopping and priority-driven strategies. This can be observed by the steep slope at the beginning of the curve of our proposed method in Fig.~\ref{fig:4}(c), during which time tasks with priority level 2 are prioritized. The high rate of tasks being aborted in N-LM is the reason why the total visited vertices for N-LM are more than for NBA-P. For N-LM, the robot will attempt to perform a task if there is still remaining resource budget. However, if the budget is exceeded during task execution, the task will be aborted and the vertex needs to be re-visited. Setting multiple priorities may be useful but this needs to be carefully tuned as having too many priority levels can make the process inefficient in practice, forcing the robot to move across the field to reach the tasks of the next highest priority level.

\begin{figure}[h]
\vspace{4pt}
    \centering
    \includegraphics[width = 0.48\linewidth]{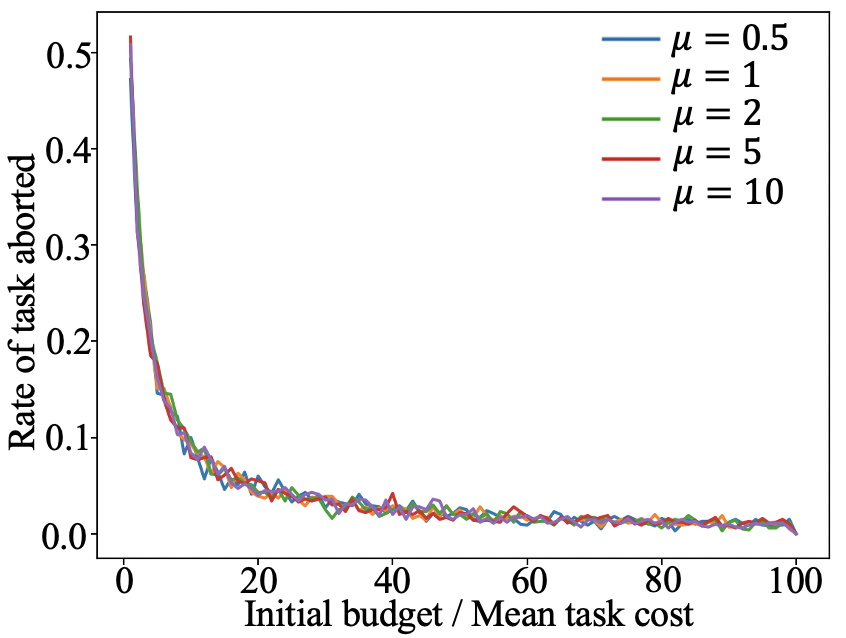}
    \includegraphics[width = 0.45\linewidth]{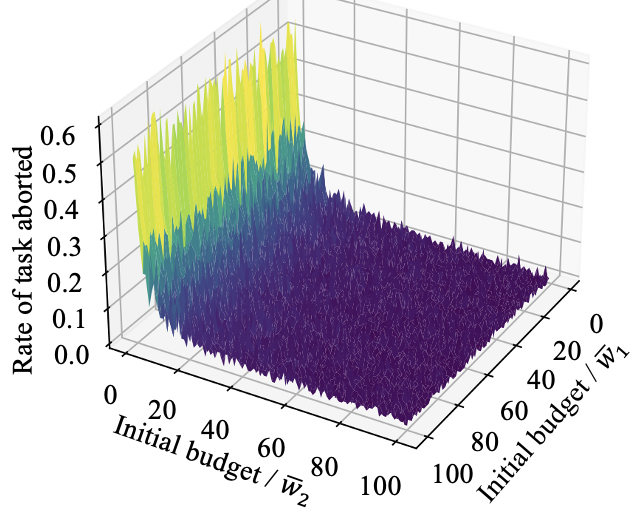}
    \vspace{-6pt}
    \caption{(Left) Rate of aborted tasks over the ratio of budget/$\bar{w}_1$ when $|S| = 1$. (Right) Rate of aborted tasks with respect to the ratio of budget/$\bar{w}_1$ and budget/$\bar{w}_2$ for the $|S| = 2$ case.}
    \label{fig:6}
    \vspace{4pt}
\end{figure}

Reducing the rate of aborted tasks can increase efficiency. We continue to study the influence of $\mu$ and $\bar{w}_s$ to the rate of aborted tasks, in which case the energy budget does not need to be considered. Starting with $|S|=1$, and assuming there exist infinite tasks, we need to determine the optimal time to stop performing more tasks. Figure~\ref{fig:6} (left) shows the relation between aborted tasks (ratio of occurrences over 1000 trials) and the ratio of initial budget over mean task cost $\bar{w}_1$, for different values of $\mu$. Results suggest that aborting a task is barely influenced by the value of $\mu$. If the initial budget is close to the mean task cost, the rate of aborted tasks can be as high as $50\%$, and the optimal stopping rule is less effective. If the initial budget is more than $50$ times the mean task cost, the rate of aborted tasks reaches $0$.

Given that $\mu$ barely affects the rate of aborted tasks (Fig.~\ref{fig:6} (left)), we examine in the case that $|S|=2$ the influence of the ratio between initial budget and task mean cost for each priority level. We assume there exist infinite tasks of priority level 1 and 2. The goal is to determine if it is better to select another task of priority level 1 or 2 to perform, or to stop. Figure~\ref{fig:6} (right) suggests that, regardless of the relation between $\bar{w}_{1}$ and $\bar{w}_{2}$, the rate of aborted tasks is more influenced by the ratio of initial budget over $\bar{w}_{2}$. This is intuitive since tasks of priority level 2 are prioritized over tasks of priority level 1. Since more $s=2$ tasks are performed if possible, it escalates its influence to the rate of aborting tasks. Thus, when higher priority tasks exist, the initial budget can be set by considering expected cost of high priority tasks, as the expected cost of low priority tasks do not have much impact when energy is sufficient. The proposed method is more suitable when the mean cost of tasks is small enough compared to the initial (resource) budget.

\begin{table*}[]
\vspace{4pt}
\centering
\caption{Results for 10 field experimental datasets in 1- and 5-robot scenarios}\label{table:2}
\vspace{-7pt}
\resizebox{1.72\columnwidth}{!}{
\begin{tabular}{|c|c|c|c|c|c|c|c|c|c|c|c|c|}
\hline
\multirow{2}{*}{\begin{tabular}[c]{@{}c@{}}robot\\ number\end{tabular}} & \multicolumn{2}{c|}{data idx} & 1 & 2 & 3 & 4 & 5 & 6 & 7 & 8 & 9 & 10 \\ \cline{2-13} 
 & \multicolumn{2}{c|}{total tasks} & \multicolumn{1}{r|}{39528} & \multicolumn{1}{r|}{44607} & \multicolumn{1}{r|}{58845} & \multicolumn{1}{r|}{38190} & \multicolumn{1}{r|}{33489} & \multicolumn{1}{r|}{24075} & \multicolumn{1}{r|}{12203} & \multicolumn{1}{r|}{58553} & \multicolumn{1}{r|}{58551} & \multicolumn{1}{r|}{20345} \\ \hline
\multirow{12}{*}{1} & \multirow{4}{*}{\begin{tabular}[c]{@{}c@{}}r/v\\ ($10^{-5}$)\end{tabular}} & NBA-P & \pmb{2.53} & \pmb{2.24} & \pmb{1.70} & \pmb{2.62} & \pmb{2.99} & \pmb{4.15} & \pmb{8.19} & \pmb{1.71} & \pmb{1.71} & \pmb{4.92} \\ \cline{3-13} 
 &  & N-LM & 1.72 & 1.23 & 0.55 & 1.23 & 1.56 & 2.17 & 6.68 & 0.59 & 0.57 & 3.72 \\\cline{3-13} 
  &  & I-LM & 2.53 & 2.24 & 1.70 & 2.61 & 2.98 & 4.14 & 8.18 & 1.70 & 1.70 & 4.91 \\\cline{3-13} 
  &  & S-GPR & 2.52 & 2.22 & 1.67 & 2.59 & 2.95 & 4.10 & 8.15 & 1.68 & 1.68 & 4.89 \\
 \cline{2-13} 
 & \multirow{4}{*}{\begin{tabular}[c]{@{}c@{}}w/v\\
 ($10^{-3}$)\end{tabular}} & NBA-P & \pmb{0.15} & \pmb{0} & \pmb{0} & \pmb{0.61} & \pmb{0} & \pmb{0} & \pmb{1.61} & \pmb{0} & \pmb{0} & \pmb{1.44} \\ \cline{3-13} 
 &  & N-LM & 5.88 & 11.9 & 23.42 & 22.07 & 16.96 & 21.37 & 6.66 & 19.45 & 24.53 & 4.91 \\\cline{3-13}
 &  & I-LM & 4.62 & 9.98 & 16.77 & 28.87 & 19.10 & 23.47 & 7.31 & 17.23 & 27.18 & 4.00 \\
 \cline{3-13} 
 &  & S-GPR & 9.12 & 22.44 & 70.44 & 49.03 & 33.47 & 41.83 & 10.94 & 58.66 & 76.18 & 7.70 \\\cline{2-13}
 & \multirow{4}{*}{\begin{tabular}[c]{@{}c@{}}visited\\ vertices\end{tabular}} & NBA-P & \pmb{39529} & \pmb{44607} & \pmb{58845} & \pmb{38192} & \pmb{33489} & \pmb{24075} & \pmb{12206} & \pmb{58553} &\pmb{58551} & \pmb{20345} \\ \cline{3-13} 
 &  & N-LM & 58220 & 81277 &181553  &81083 & 64178 & 46144 & 14967 &168360 & 176174 & 26914 \\\cline{3-13} 
 &  & I-LM & 39585 & 44703 & 58966  & 38349 & 33600 & 24159 & 12223 & 58689 & 58734 & 20372 \\\cline{3-13} 
 &  & S-GPR & 39731 & 45018 & 59933 & 38672 & 33847 & 24363 & 12263 & 59504 & 59587 & 20433 \\ \hline
\multirow{12}{*}{5} & \multirow{4}{*}{\begin{tabular}[c]{@{}c@{}}r/v\\ ($10^{-5}$)\end{tabular}} & NBA-P &  \pmb{2.53} & \pmb{2.24} & \pmb{1.70} & \pmb{2.62} & \pmb{2.99} & \pmb{4.15} & \pmb{8.19} & \pmb{1.71} & \pmb{1.71} & \pmb{4.91} \\ \cline{3-13} 
 &  & N-LM & 1.77 & 1.3 & 0.62 & 1.37 & 1.69 & 2.33 & 6.81 & 6.52 & 6.03 &  3.81 \\\cline{3-13} 
 &  & I-LM & 2.53 & 2.24 & 1.70 & 2.61 & 2.98 & 4.14 & 8.18 & 1.70 & 1.70 & 4.91 \\\cline{3-13} 
 &  & S-GPR & 2.52 & 2.22 & 1.67 & 2.59 & 2.95 & 4.10 & 8.15 & 1.68 & 1.68 & 4.89 \\ \cline{2-13} 
 & \multirow{4}{*}{\begin{tabular}[c]{@{}c@{}}w/v\\ ($10^{-3}$)\end{tabular}} & NBA-P & \pmb{0} & \pmb{0} & \pmb{0} & \pmb{0.92} & \pmb{0} & \pmb{0} & \pmb{2.67} & \pmb{0} & \pmb{0} & \pmb{0.59} \\ \cline{3-13} 
 &  & N-LM & 5.97 & 12.00 & 24.72 & 24.67 & 18.66 & 22.62 & 6.23 & 21.18 & 27.05 &  4.72\\\cline{3-13}
 &  & I-LM & 5.81 & 8.37 & 14.78 & 27.35 & 18.47 & 23.58 & 4.94 & 14.66 & 29.46 & 3.61 \\\cline{3-13}
 &  & S-GPR & 9.08 & 22.66 & 70.37 & 49.03 & 33.09 & 41.31 & 10.26 & 58.66 & 76.18 & 7.79 \\ \cline{2-13} 
 & \multirow{4}{*}{\begin{tabular}[c]{@{}c@{}}visited\\ vertices\end{tabular}} & NBA-P & \pmb{39528} & \pmb{44607} & \pmb{58845} & \pmb{38193} & \pmb{33489} & \pmb{24075} & \pmb{12209} & \pmb{58553} & \pmb{58551} & \pmb{20348} \\ \cline{3-13} 
 &  & N-LM & 56543 & 77210 & 162420 & 73179 & 59253 & 42923 & 14688 & 153425 & 165869 & 26233 \\\cline{3-13} 
 &  & I-LM & 39599 & 44687 & 58952 & 38335 & 33596 & 24160 & 12218 & 58670 & 58753 & 20369 \\
 \cline{3-13} 
 &  & S-GPR & 39729 & 45020 & 59933 & 38672 & 33847 & 24363 & 12263 & 59504 & 59587 & 20433 \\
 \hline
\end{tabular}}
\vspace{-18pt}
\end{table*}

\subsection{Testing with Real-world Field Data}

The real-world datasets used here contain soil moisture values collected in a commercial vineyard located in central California. The structure of the vineyard imposes motion constraints to ground robots moving therein. Irrigation lines are attached to metallic wires at about $12$\;in from the ground and running parallel to the trellis. Thus, to move from one row to another (even if adjacent), the robot must first exit the row from either end (based on its direction), and then re-enter at the desired row. Samples were collected on a regular grid with 275 rows and 214 columns uniformly covering the field. Sampling locations were computed offline, and data were collected with a Campbell H2SP soil moisture sensor.

\begin{figure}[h]
\vspace{-9pt}
    \centering
    \includegraphics[trim={0 0.5cm 0 7.5cm},clip,height=0.35\linewidth]{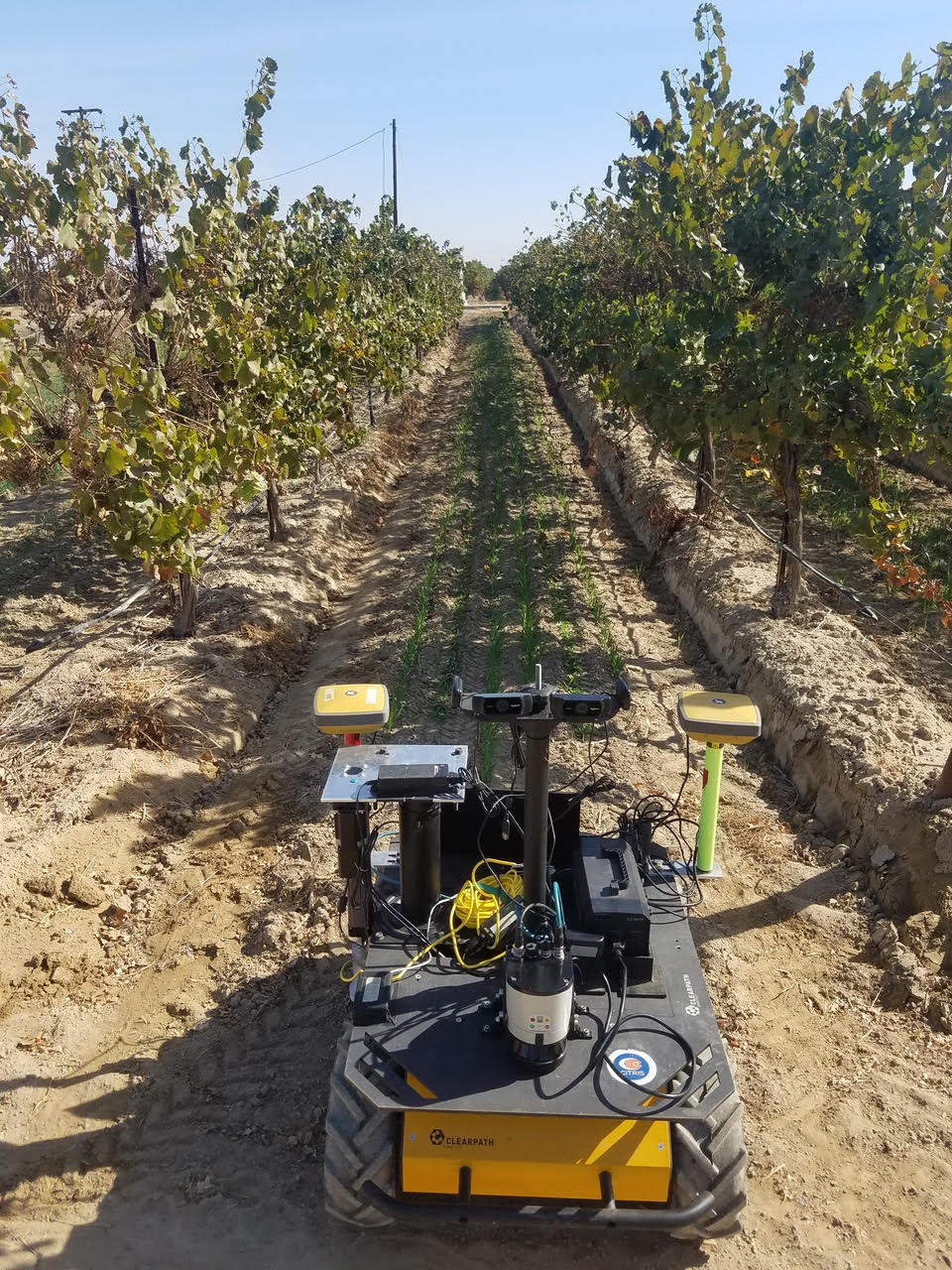}
     \includegraphics[trim={2.5cm 0.6cm 2.5cm 1.2cm},clip,height=0.35\linewidth]{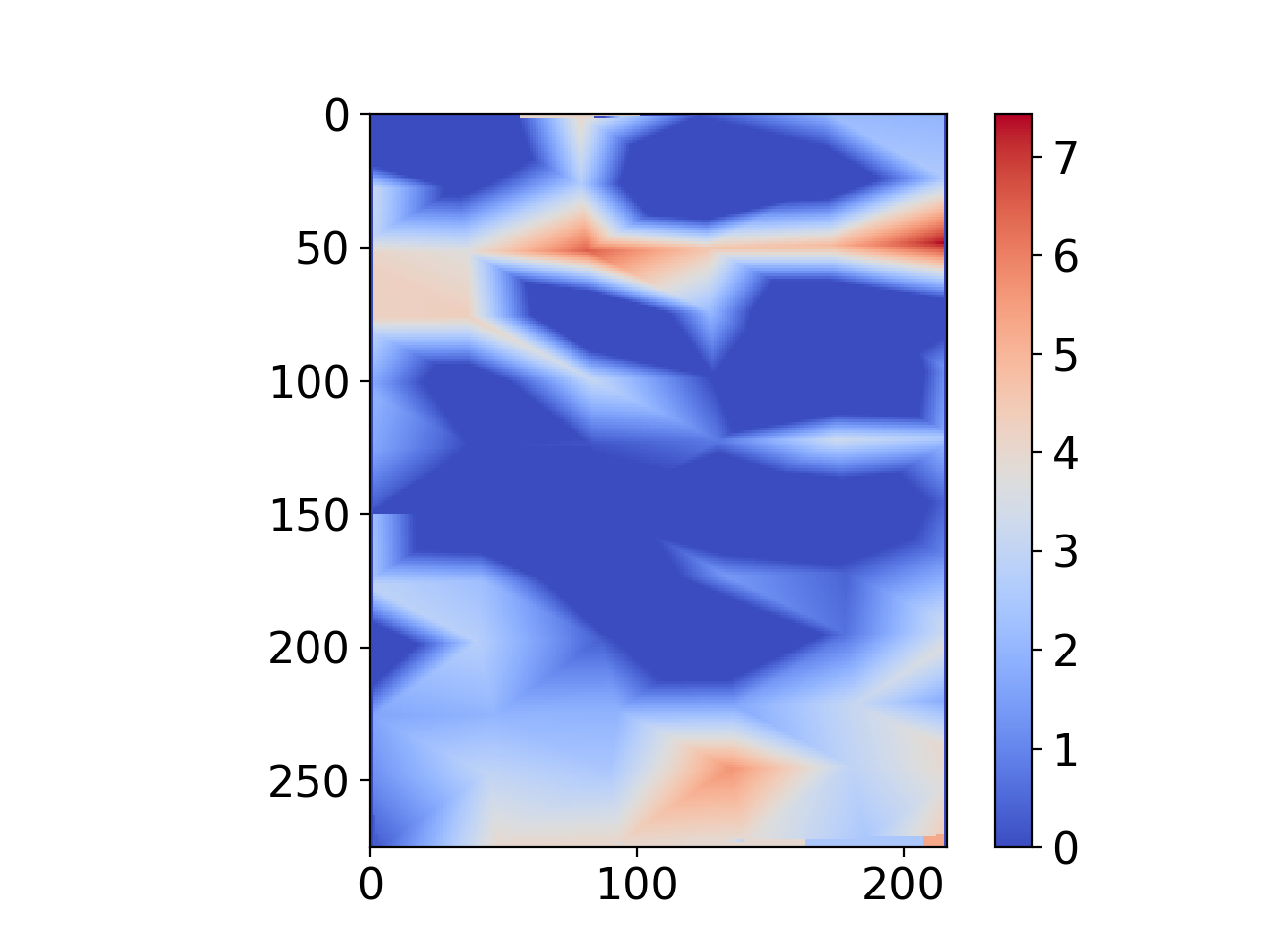}       
    \vspace{-6pt}
    \caption{(Left) Instance of a wheeled mobile robot performing soil moisture measurements in a commercial vineyard. (Right) Sample ground truth cost (i.e. the moisture difference between collected moisture values and a desired level) for one of the field experimental datasets used here. Low-moisture (dry) locations are indicated by higher differences. In these areas, more water (the resource in this case) needs to be consumed to reach a desired moisture level, which is equivalent to a higher cost. Discretely-sampled values where mapped to a continuous contour illustrated here using the kriging interpolation algorithm. (However, we use the discrete values directly on the SAG representation of the environment utilized by our algorithm.)}
    \label{fig:7}
    \vspace{-6pt}
\end{figure}

Suppose autonomous ground mobile robots are deployed to water plants in the vineyard. Vertices with moisture values less than a desired level are considered to contain a task (of precise watering). The ground truth task cost at any vertex is the moisture difference between collected moisture values and the desired level. An example in shown in Fig.~\ref{fig:7}. The robots' decision is constrained by the resource budget of total water carrying capacity, and the energy budget to move between locations. All tasks are considered to have the same priority level, i.e. $|S|=1$. The location of tasks and the mean cost of all tasks (averaged real costs of all tasks using ground truth) are available to the robot(s) prior to departure, whereas the actual cost for each task is unknown to the robot(s) before task completion. Without loss of generality, we consider the movement cost on edges to be 1 (all water emitters are located within the same interval distance). We test in 1- and 5-robot scenarios, where the robots depart with energy budget $800$ and resource budget $400$. The base stations are located at $v_{137,0}$ and $v_{137,215}$. 

Table~\ref{table:2} shows results for 10 real-world datasets in 1- and 5-robot scenarios. The 10 datasets were collected during a timespan of 5 months, hence task locations and costs differ among datasets. Results suggest that, in all cases, our proposed NBA-P algorithm achieves higher $r/v$ ratio, lower $w/v$ ratio, and less total visited vertices than N-LM. Even though I-LM and S-GPR achieve similar $r/v$ ratio and total visited vertices as NBA-P, the $w/v$ ratio is much higher compared to NBA-P. Results  attempts only if the expected task cost is less than the remaining budget yet it fails to consider the uncertainties in actual resource consumption, which can be much higher than the expected value. In addition, the high actual cost may cause multiple failed attempts at the same position. Thus, the total waste of resource can be significant, evident by the averaged total waste over 10 datasets for 1-robot cases being 5, 1725, 671 and 1796 for NBA-P, N-LM, I-LM and S-GPR, respectively. That is, NBA-P is able to handle uncertain task costs, and requires less total resources to complete the same amount of tasks as compared to other methods. Importantly, no tasks are aborted in 13 out of 20 cases using NBA-P, where each case contains up to 60000 tasks. For datasets 3, 8, and 9, N-LM in fact visits three times more vertices than the proposed method. The total path lengths for evaluated methods differ within around $2\%$-range, and NBA-P achieves the shortest path for datasets 6 and 10. That is, NBA-P plans paths of similar length as lawnmower methods.

In all, testing with experimental data validates the efficacy and efficiency of our proposed method, and demonstrates preliminary feasibility that it can scale both in terms of the size of the environment and the number of robots in the team.

\section{Conclusions}
\paragraph*{\textbf{Contributions and Key Findings}} The paper contributes to stochastic task allocation in precision agriculture. Given resource and energy budgets, our NBA-P algorithm returns the best action on a stochastic aisle graph (SAG) by simultaneously determining optimal sampling locations and optimal stopping times to return to a base station, all at run-time. The proposed algorithm is tested using both simulated data for a 2-robot scenario and agricultural field experimental datasets for 1- and 5-robot scenarios. Results suggest that, by applying NBA-P, all tasks are completed, and tasks with high priority levels are prioritized when possible. The rate of aborted tasks is minimal when the initial resource budget is more than 50 times the mean task costs. NBA-P outperforms N-LM, I-LM and S-GRP methods in all simulated and real-world datasets, in terms of more gain per vertex visited, fewer tasks being aborted, and less total visited vertices to complete the same number of tasks. In testing with real-world datasets, our method has no tasks aborted in 13 out of 20 cases with up to 60000 tasks in each case. Further, N-LM visits up to 3 times more vertices than NBA-P to complete same number of tasks, which leads to a significant waste of resources.

\paragraph*{\textbf{Directions for Future Work}} At its current form, NBA-P is not suitable for scenarios where the task and movement costs are correlated. Further, the overall paths using NBA-P can be longer than those derived via the lawnmower method, especially when multiple priority levels exist and tasks of different priority level are intertwined.  
Future directions of research include 1) application of the proposed algorithm to physical robots in the field, and 2) study of the scenario that considers correlated cost for movement and task execution. 

\appendix
Let \textit{hl} and \textit{hr} represent the robot moving toward column $0$ and $j+1$, respectively.
Suppose $i_{11} = min(i_c,i')$, $i_{12} = max(i_c,i')$, $i_{21} = min(i',i_d)$, and $i_{22} = max(i',i_d)$. Then,
\begin{equation*}
\begin{medsize}
\begin{split}
    t_{\alpha}(i') &=
    \begin{cases}
    \sum_{j=j_c}^{n-1}c_e(e_{i_c,j^+}) + \sum_{i=i_{11}}^{i_{12}-1}c_e(e_{i^+,n+1}), \text{if } \textit{hl},\\
    \sum_{j=1}^{j_c-1}c_e(e_{i_c,j^+}) + \sum_{p=i_{11}}^{i_{12}-1}c_v(e_{i^+,0}), \text{if } \textit{hr}.
    \end{cases}\\
    t_{\beta}(i') &=  \sum_{j=1}^{n-1}c_e(e_{i',j^+}).\\
    t_{\gamma}(i') &=
        \begin{cases}
        \sum_{i=i_{21}}^{i_{22}-1} c_e(e_{i^+,0}), \text{if } \textit{hl},\\ \sum_{i=i_{21}}^{i_{22}-1} c_e(e_{i^+,n+1}), \text{if } \textit{hr}.
        \end{cases}
\end{split}
\end{medsize}
\end{equation*}

\bibliographystyle{IEEEtran}
\bibliography{IEEEabrv,IEEEexample}

\end{document}